\documentclass[10pt,twocolumn,letterpaper]{article}

\usepackage{cvpr}      
\usepackage{booktabs}
\usepackage{multirow}
\usepackage[table]{xcolor}
\usepackage{amsmath}
\usepackage{makecell}
\usepackage{algorithm}
\usepackage{algorithmicx}
\usepackage{float}
\usepackage{algpseudocode}

\usepackage{tikz} 
\usepackage[table]{xcolor}
\usepackage{booktabs,multirow,makecell}
\definecolor{accent}{HTML}{C2410C} 
\newcommand{\best}[1]{\cellcolor{green!22}\textbf{#1}}
\newcommand{\midc}[1]{\cellcolor{yellow!25}#1}
\newcommand{\worst}[1]{\cellcolor{orange!20}#1}
\makeatletter
\renewcommand\paragraph{\@startsection{paragraph}{4}{\z@}%
  {-1.0ex plus -0.2ex minus -0.2ex}
  {-1em}
  {\normalfont\normalsize\bfseries}}
\makeatother

\definecolor{cvprblue}{rgb}{0.21,0.49,0.74}
\usepackage[pagebackref,breaklinks,colorlinks,allcolors=cvprblue]{hyperref}

\usepackage{xcolor}

\makeatletter
\@ifundefined{todo}{
  
}{
  
}
\makeatother

\def\paperID{*****} 
\def\confName{CVPR}
\def\confYear{2026}

\title{\vspace{-1.5em}BAgger: Backwards Aggregation for \\Mitigating Drift in Autoregressive Video Diffusion Models}

\author{
Ryan Po \quad
Eric Ryan Chan \quad
Changan Chen \quad
Gordon Wetzstein \\[0.6em]
Stanford University \\[0.6em]
\url{https://ryanpo.com/bagger}
\vspace{-0.2em}
}

\begin{document}
\twocolumn[{%
\renewcommand\twocolumn[1][]{#1}%
\maketitle
\begin{center}
    \vspace{-2em}
    \centering
    \captionsetup{type=figure}
    \includegraphics[width=\linewidth]{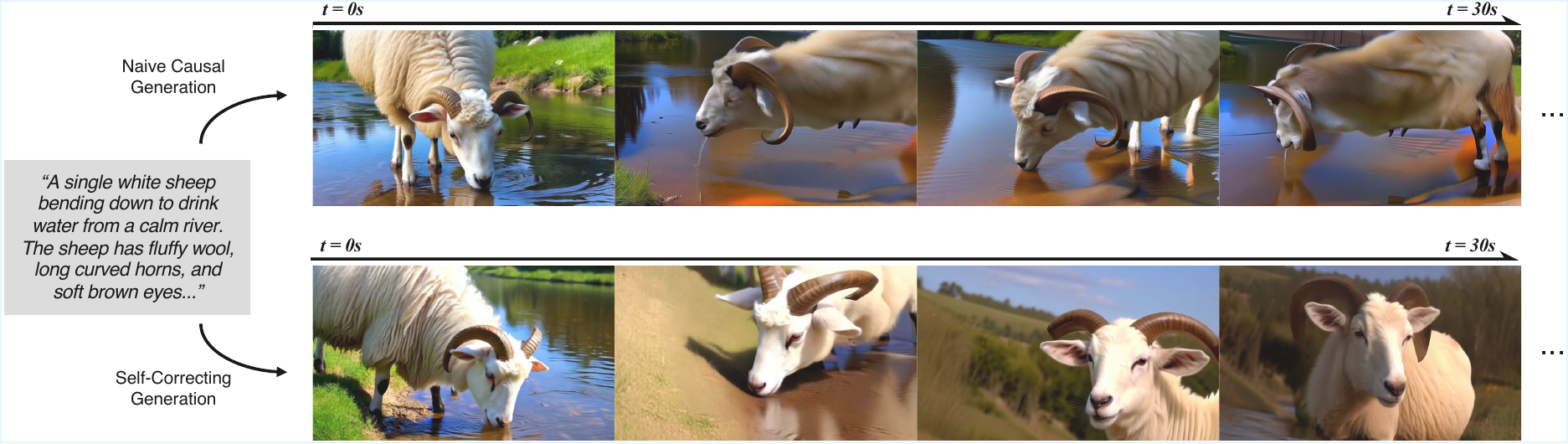}
    \vspace{-2em}
    \captionof{figure}{
   In standard autoregressive video diffusion, exposure bias causes compounding errors (``drift'') as generated frames become the context for future steps. This is typically observed as degrading image quality, contrast, and disintegration of content (top row). We reverse the model’s own rollouts to form corrective trajectories that exemplify how to undo drifted states. A DAgger-style aggregation loop fine-tunes the model on these corrective samples using the original diffusion objective—no bi-directional teacher or distribution matching objectives required. The resulting model exhibits more stable, diverse long-horizon generations (bottom row).
    }
    \label{fig:teaser}
\end{center}%
}]
\begin{abstract}
Autoregressive video models are promising for world modeling via next-frame prediction, but they suffer from exposure bias: a mismatch between training on clean contexts and inference on self-generated frames, causing errors to compound and quality to drift over time. We introduce Backwards Aggregation (BAgger), a self-supervised scheme that constructs corrective trajectories from the model’s own rollouts, teaching it to recover from its mistakes. Unlike prior approaches that rely on few-step distillation and distribution-matching losses, which can hurt quality and diversity, BAgger trains with standard score or flow matching objectives, avoiding large teachers and long-chain backpropagation through time. We instantiate BAgger on causal diffusion transformers and evaluate on text-to-video, video extension, and multi-prompt generation, observing more stable long-horizon motion and better visual consistency with reduced drift.
\vspace{-1em}
\end{abstract}    

\section{Introduction}
\label{sec:intro}
Video world models are causal generative models designed to predict how the visual world evolves given a set of actions or prompts~\cite{Bruce2024GenieGI,parkerholder2024genie2,genie3, Agarwal2025CosmosWF, nvidia2025worldsimulationvideofoundation, Li2025HunyuanGameCraftHI, Ha2018WorldM, Po2025LongContextSV, Wu2025VideoWM}. Recently, video diffusion models have proven to be a promising approach for world modeling~\cite{huang2025towards}. While earlier video diffusion models focus on generating fixed-length clips in a non-causal, i.e., bi-directional, manner~\cite{Yang2024CogVideoXTD,Wang2025WanOA}, the task of world modeling inherently requires causality, leading to a shift in adoption of autoregressive architectures~\cite{ai2025MAGI1AV, Ge2022LongVG, Hong2022CogVideoLP, Kondratyuk2023VideoPoetAL, Yan2021VideoGPTVG, Yu2023LanguageMB}. 

However, common to all autoregressive generative models, causal video diffusion suffers from exposure bias~\cite{Ning2023ElucidatingTE, Schmidt2019GeneralizationIG}, an effect that leads to rapid deterioration of video quality over time~\cite{Wang2025ErrorAO, Yin2024FromSB, Huang2025SelfFB, Zhang2025FrameCP}. Exposure bias, also known as drift, stems from a fundamental mismatch between the training and inference objectives. During training, the model learns to condition on \textit{ground truth} context frames from the previous step, providing clean and reliable context for the next step of generation. In contrast, during inference the model is given \textit{generated} frames from the previous step to generate the next video frames. If the model makes an error during generation, these errors are compounded into subsequent steps, leading to a cascading effect that degrades video quality over time~\cite{Yin2024FromSB}.

\begin{figure*}[t]
    \centering
    \includegraphics[width=\linewidth]{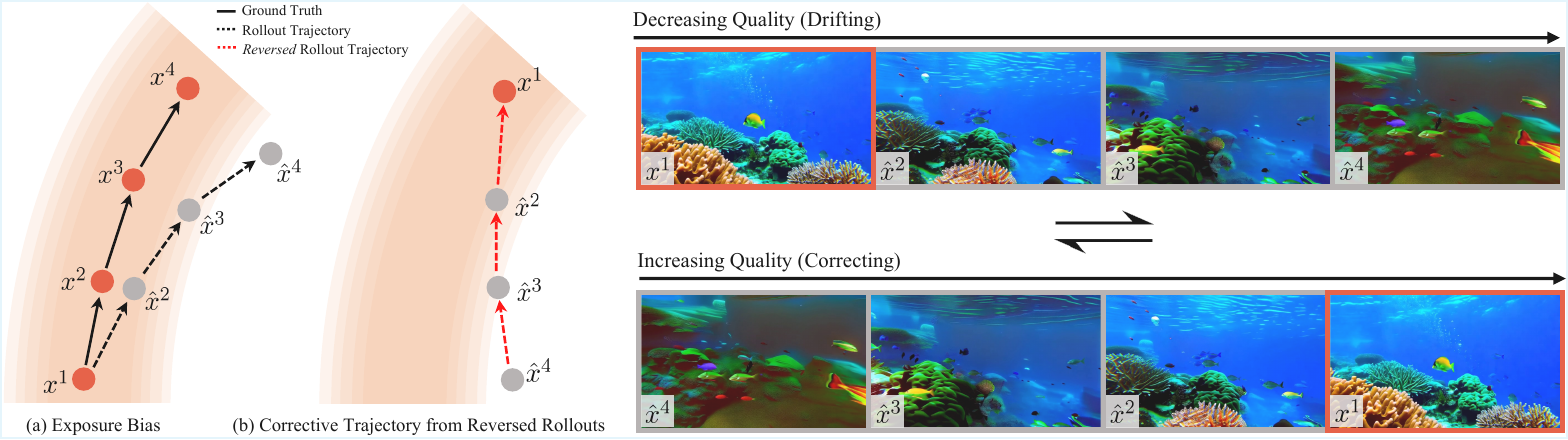}
    \vspace{-1.8em}
    \caption{\textbf{Backwards Model Rollouts as Corrective Trajectories.} (a) Exposure bias arises from a mismatch between training and inference objectives. During inference, small errors accumulate and cause generations $\hat{x}^{2:4}$ to drift from the real distribution $x^{2:4}$. The orange contours depict regions of higher probability under the video data distribution.
    (b) Given an AR rollout that may exhibit drift, we treat the reversed clip as a supervision signal for recovering from drifted contexts. This converts the model’s own mistakes into training data without requiring an external expert or teacher network.}
    \vspace{-1em}
    \label{fig:corrective_traj}
\end{figure*}

Prior works that tackle exposure bias usually aim at closing the training--inference gap by aligning the generated distribution from the autoregressive model with a pre-trained bidirectional teacher~\cite{Huang2025SelfFB, Cui2025SelfForcingTM, Liu2025RollingFA, Yang2025LongLiveRI, Shin2025MotionStreamRV}. However, this imposes three key limitations: (1) the reliance on a pre-trained bi-directional teacher diffusion model~\cite{Wang2025WanOA}, (2) the need for back-propagation through time (BPTT) over the entire autoregressive generation process, and (3) the distribution-matching loss itself~\cite{Treat2018GENERATIVEAN,Yin2023OneStepDW, Yin2024ImprovedDM}, which is known to be mode seeking and can suppress diversity. Other methods also seek to mitigate drift by injecting noise into context frames~\cite{Zhang2025FrameCP, Chen2024DiffusionFN, Song2025HistoryGuidedVD, Valevski2024DiffusionMA, oasis2024}, but this often degrades temporal consistency and fails to address exposure bias at its core.


\begin{figure*}[t]
    \centering
    \includegraphics[width=\linewidth]
    {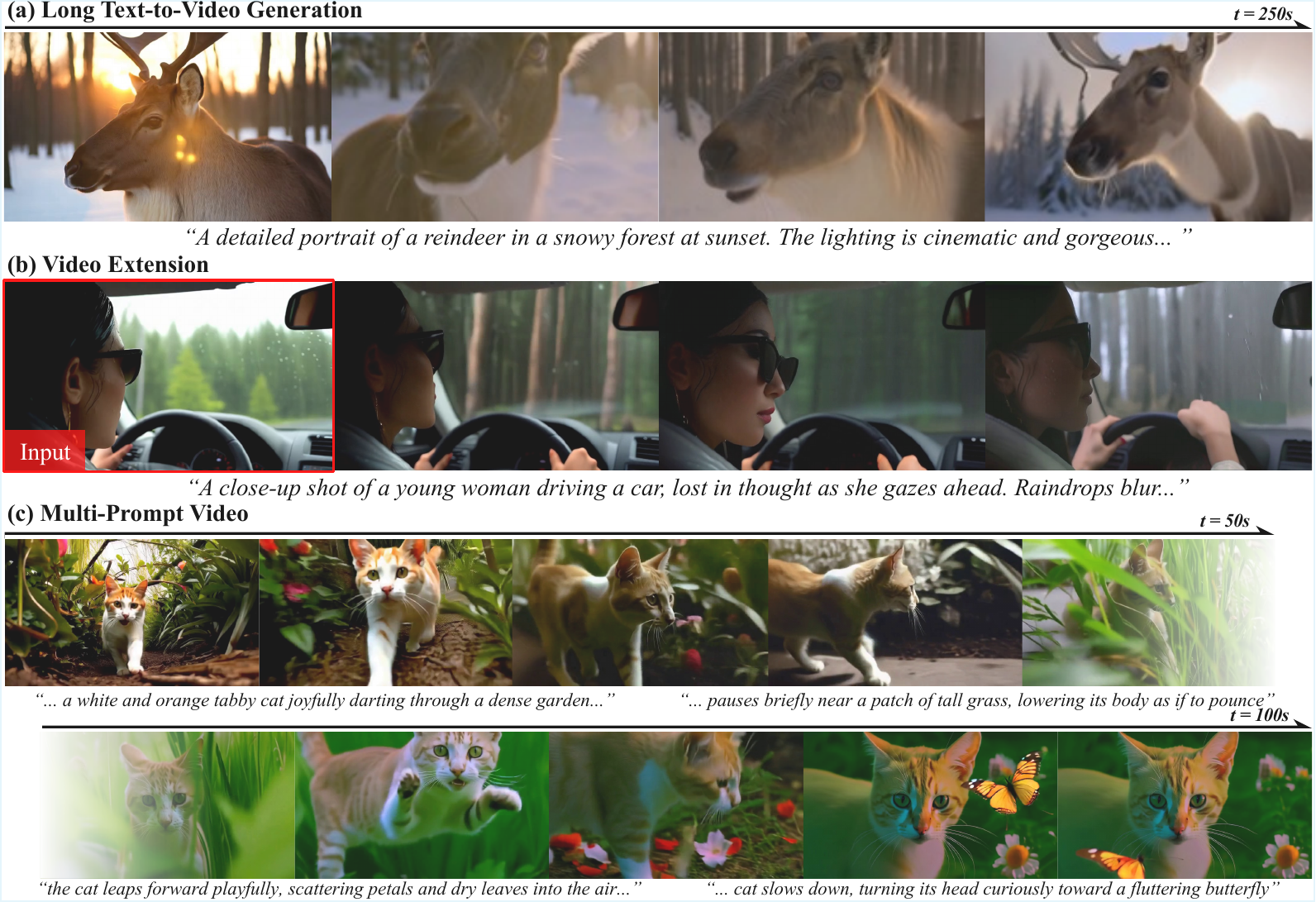}
    \vspace{-1.8em}
    \caption{\textbf{Applications of BAgger Trained AR Video Model}. (a) Text-to-video generation from a static text prompt, producing a 30-second clip that maintains subject identity and cinematic appearance over time. (b) Video extension, where our model continues an input clip while preserving the input’s appearance and motion dynamics. (c) Multi-prompt generation, in which the text conditioning is updated over time, enabling fine-grained, temporally coherent control over the evolving video.}
    \label{fig:main_results}
    \vspace{-1.3em}
\end{figure*}

We propose a method for overcoming drift by guiding a model to learn how to correct its own mistakes. Drawing inspiration from Dataset Aggregation (DAgger)~\cite{Ross2010ARO}, we iteratively construct and fine-tune a model on a dataset of ``corrective trajectories''. This teaches the model to be robust to its own mistakes, correcting correcting errors made during generation instead of allowing them to compound to subsequent steps. In the field of imitation learning, where DAgger is often used to great effect, these ``corrective trajectories'' are gathered through expert intervention, where a human manually corrects mistakes made by the trained policy. 
However, in the case of video generation, ``corrective trajectories'', i.e. smooth, continuous transitions from low-quality clips to high-quality clips, cannot be obtained via human intervention. 
In our work, we leverage the following observation: by reversing a model's own generated rollouts, which start from high-quality real videos and degrade autoregressively in time, we are naturally provided with examples of how the model can correct its own errors. This is uniquely supported by the fact that time-reversed videos remain in-distribution for text-to-video models as long as the prompt is updated to describe the reversed motion.


We show that by repeatedly aggregating more ``corrective'' samples, the resulting model learns to correct errors made during its own generation process, leading to better generation quality over long horizons compared to teacher/diffusion-forcing training. We also identify how the characteristics of errors change across multiple rounds of training, and propose a method for modifying the training regime, preventing quality degradation due to training on generated data. We compare our method against prior works performing autoregressive video generation, showing qualitative and quantitative improvements. 
We summarize our contributions as follows:
\begin{itemize}
    \item We introduce \textit{\textbf{Backwards Aggregation (BAgger)}}, which leverages reversed model rollouts as self-corrective signals to enhance stability over long-horizon generations.
    \item Inspired by Dataset Aggregation~\cite{Ross2010ARO}, we iteratively construct and train on a dataset of ``corrective trajectories'', gradually teaching a model to learn stable generative trajectories while conditioned on drifted states.
    \item We evaluate our model on diverse tasks—including text-to-video, video extension, and multi-prompt generation—and show improved long-horizon performance in both qualitative and quantitative evaluations.
\end{itemize}

\section{Related Work}
\label{sec:related_work}
\paragraph{Video Diffusion Models.} 
Current state-of-the-art in video generation models have mostly been set by large-scale bidirectional diffusion transformers~\cite{Peebles2022ScalableDM, videoworldsimulators2024}, demonstrating remarkable capabilities in synthesizing high-quality, complex video clips. The core mechanism relies on full spatiotemporal attention applied to all tokens~\cite{videoworldsimulators2024}, as every video token is denoised simultaneously~\cite{Blattmann2023StableVD, Blattmann2023AlignYL, videoworldsimulators2024, Li2024AutoregressiveIG, Gupta2023PhotorealisticVG, HaCohen2024LTXVideoRV, Ho2022ImagenVH, Ho2022VideoDM, Kong2024HunyuanVideoAS, Polyak2024MovieGA, Villegas2022PhenakiVL, Wang2025WanOA, Hong2022CogVideoLP, Yang2024CogVideoXTD, BarTal2024LumiereAS, Cai2025MixtureOC}, generating videos of fixed lengths. However, because these models are inherently non-causal and constrained to fixed-length generation, they are not well suited to serve as true world models.

\paragraph{Autoregressive Video Models.} Autoregressive approaches in video generation factorizes the joint distribution over all frames into, $p(x^{1:N}) = \prod^N_{i=1} p(x^i | x^{<i})$. This formulation naturally aligns with the causality of time, as video frames are generated sequentially, making autoregressive models well suited for interactive simulation~\cite{Shin2025MotionStreamRV, Huang2025SelfFB} and world modeling tasks~\cite{genie3,Li2025HunyuanGameCraftHI, Wu2025VideoWM}. Conventional autoregressive video models rely on direct next-token prediction of discrete video tokens~\cite{Bruce2024GenieGI, Kondratyuk2023VideoPoetAL, Ren2025NextBP, Wang2024LoongGM, Weissenborn2019ScalingAV, Yan2021VideoGPTVG}. However, the performance of discretized AR models often lag behind diffusion models. To address this, recent works have explored training strategies combining autoregression and diffusion~\cite{Gao2024Ca2VDMEA, Gu2025LongContextAV, Guo2025LongCT, Hu2024ACDiTIA, Jin2024PyramidalFM, Li2024ARLONBD, Liu2024MarDiniMA, Liu2024RedefiningTM, Weng2023ARTVAT, Zhang2025TestTimeTD, Zhang2025GenerativePA}. Some works train conditional diffusion models that denoise next frames condition on past clean frames~\cite{Zhang2025FrameCP, Huang2025SelfFB}, while other approaches introduce per-frame independent noise levels during training~\cite{Chen2024DiffusionFN, Song2025HistoryGuidedVD, Valevski2024DiffusionMA}, allowing for AR inference.


\begin{figure*}[t]
    \centering
    \includegraphics[width=\linewidth]{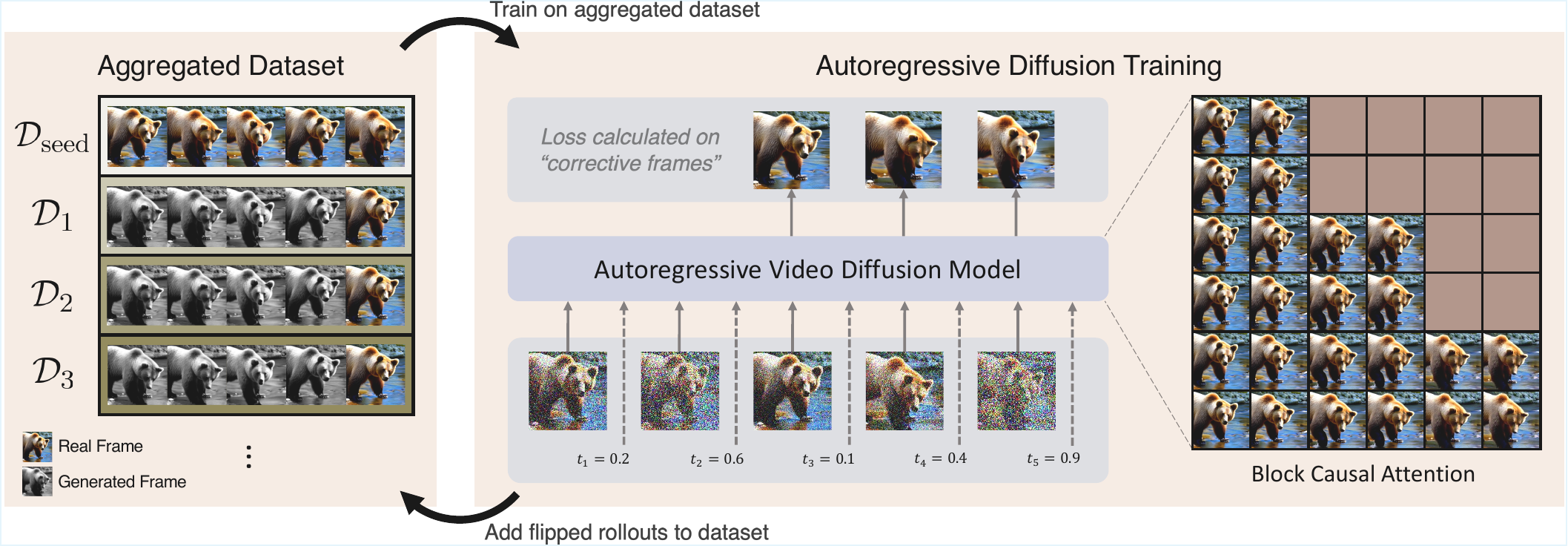}
    \vspace{-1.8em}
    \caption{\textbf{BAgger Training Loop.} Starting from an initial dataset of ground truth video data $\mathcal{D}_{\text{seed}}$, each round: (1) train a Diffusion Forcing model (policy) on the dataset; (2) sample on-policy rollouts starting from a ground truth frame; (3) reverse them to form corrective trajectories; (4) aggregate with the dataset; (5) continue training with the original diffusion objective. Iterating this process closes the train–test gap by teaching the model to act on its own drifted states.}
    \label{fig:placeholder_doublecol}
    \vspace{-1.5em}
\end{figure*}

\paragraph{Mitigating Exposure Bias.} Several training strategies have been proposed to mitigate exposure bias. Diffusion Forcing (DF)~\cite{Chen2024DiffusionFN} trains a diffusion model with varying per-frame independent noise levels. With DF, adding noise to context frames during training has shown to mitigate drift~\cite{Valevski2024DiffusionMA}, as the model becomes more robust to perturbations in the context. However, DF models still suffer from error accumulation, since the noised context frames remain mismatched from the inference-time context distribution~\cite{Yin2024FromSB, Valevski2024DiffusionMA}. History Guidance~\cite{Song2025HistoryGuidedVD} proposes a conditioning mechanism that guides the generation process during inference to remain consistent with its past. Although effective along short horizons, History Guidance can worsen error accumulation, as the model is guided to stay consistent even with drifted frames. Self Forcing~\cite{Huang2025SelfFB} takes a more principled approach towards closing the train--test gap by performing autoregressive self-rollout during training. Aligning the distribution of its own rollouts against the distribution of a pre-trained bidirectional teacher, the resulting model is robust to error accumulation for dozens of seconds, with follow-up works extending this to minutes~\cite{Yang2025LongLiveRI, Cui2025SelfForcingTM, Liu2025RollingFA}. While effective, this approach fundamentally relies on distribution-matching losses such as score distillation~\cite{Yin2024ImprovedDM, Yin2023OneStepDW} or adversarial objectives~\cite{Treat2018GENERATIVEAN}, which are known to be less stable and accurate for density estimation compared to flow or score-matching objectives~\cite{Lipman2022FlowMF}. 
SVI~\cite{Li2025StableVI} explores this problem through noise injection, but only for the non-causal image-conditioned setting.


\paragraph{Dataset Aggregation.} Our approach is fundamentally inspired by Dataset Aggregation (DAgger)~\cite{Ross2010ARO}, an established algorithm from the field of imitation learning for addressing exposure bias. Originally designed to mitigate distributional shift in behavioral cloning, DAgger teaches a model to ``learn from its own mistakes'' through an iterative process. A single round of DAgger, (1) trains a policy with an initial dataset of observations, (2) rolls out the model to collect on-policy states, (3) queries an expert for correct actions in those states, and (4) aggregates the new state-action pairs with the existing dataset to re-train the policy. Our method follows a similar approach, with the key insight that reversing an AR generated video can be viewed as an oracle-free method for obtaining ``corrective'' actions. Over many such cycles, DAgger progressively builds a dataset that approximates the model’s own trajectory distribution in the limit, effectively resolving the train–test mismatch at the heart of the exposure bias problem.

\section{Method}
\label{sec:methods}

\subsection{Preliminary: Autoregressive Video Diffusion}

Autoregressive video diffusion models combine autoregressive factorization with denoising diffusion models, leveraging the benefits of both~\cite{Chen2024DiffusionFN}. Specifically, they make the factorization $p(x^{1:N}) = \prod^N_{i=1} p(x^i | x^{<i})$, and model the conditional distribution of each frame $p(x^i | x^{<i})$ as a diffusion process. This requires training a denoiser network, $\hat{\epsilon}_\theta$ to predict and remove noise $\epsilon^i$ from a corrupted frame $x_t^i$ at timestep $t$. The distinction from non-causal models is that this prediction is conditioned on preceding context frames $x^{<i}$. During training, this context can be kept clean (Teacher Forcing) or have independently sampled noise injected (Diffusion Forcing). Specifically, Diffusion Forcing works by sampling noise levels independently for each frame in a training sequence. A ground truth sequence $x_0^{1:N}$ is corrupted such that each from $x_0^i$ is noised to a different, randomly sampled timestep $t_i$:
\begin{equation}
    x^i_{t_i} = \alpha_{t_i}x^i_0 + \sigma_{t_i}\epsilon^i.
\end{equation}
The denoiser $\hat{\epsilon}_{\theta}$ is then trained to reverse the diffusion process, predicting the noise $e^i$ conditioned on all previous noisy context frames $x_{t_j}^{j<i}$. The model is trained on a frame-wise denoising objective:
\begin{equation}
\label{eq:diffusion_forcing}
    \mathcal{L}_{DF}(\theta) = \mathbb{E}_{t_i, x_0, \epsilon^i} \left[ ||\hat{\epsilon}_{\theta}(x_{t_i}^{i}, t_i, x_{t_j}^{j<i}) - \epsilon^i||_{2}^{2} \right]
\end{equation}
The above training objective naturally enables autoregressive inference, as denoising a noisy frame $x_t^i$ given clean context frames $x_0^{<i}$ is simply a special case covered by the training distribution. 

In this work, we train our autoregressive models using the Diffusion Forcing~\cite{Chen2024DiffusionFN} objective. We primarily focus on using a transformer-based architecture~\cite{Peebles2022ScalableDM} and enable autoregressive factorization through the use of causal attention~\cite{ai2025MAGI1AV, Yin2024FromSB, Po2025LongContextSV, Huang2025SelfFB}, allowing consistent video generation capabilities across extended temporal horizons.

\subsection{Backwards Aggregation (BAgger)}

\paragraph{Learning Corrective Trajectories.} 
As previously discussed, exposure bias originates from a fundamental mismatch between the model's training and inference-time distributions. During training, the model is exposed exclusively to ground-truth data. It learns to model the distribution $p(x^i | x^{<i})$, where $x^{<i}$ is always drawn from the real data manifold. During inference, the model must sample from $p(x^i | \hat{x}^{<i})$, conditioned on its own imperfect, self-generated outputs $\hat{x}^{<i}$. 

To close this train--test gap, the model must be trained to be robust to its own imperfect outputs, learning to correctly model the test-time distribution $p(x^i | \hat{x}^{<i})$. While autoregressive rollout provides the drifted context states $\hat{x}^{<i}$, collecting the corrective examples to train the desired inference time policy $p(x^i|\hat{x}^{<i})$ is non-trivial. In classic behavioral cloning settings~\cite{Torabi2018BehavioralCF}, the DAgger algorithm collects these examples by querying an expert oracle. For example, in autonomous driving~\cite{nvidia2025worldsimulationvideofoundation, Agarwal2025CosmosWF, Gao2025FoundationMI, Chen2023EndtoEndAD}, if the policy drifts away from the center of the lane, a human expert can intervene and provide a specific corrective action (e.g. steering towards the center) for the drifted state. However, this data collection process is not directly applicable to the setting of AR video generation. Error accumulation in videos is not a low-dimensional, discrete error, but a subtle, high-dimensional degradation in the latent space, manifested in forms such as progressive over-saturation, over-smoothing, or loss in motion diversity~\cite{Yin2024FromSB}. A human expert cannot feasibly intervene and manually create a set of ``corrective'' frames that reverses the compounded errors. Therefore, we must rely on a different source to provide expert supervision needed to generated these corrective trajectories.

\paragraph{Reversed Rollouts as Corrective Trajectories.} We make the observation that reversing an AR video model's own rollouts provides effective corrective trajectories for training a model that is robust to its own mistakes. Specifically, given some initial ground truth frame $x^1 \sim p_{\text{data}}(x)$, we autoregressively sample a video 
\begin{equation}
    (x^1, \hat{x}^{2:N}) \sim p_\theta(x^{2:N} | x^1, c) = \prod^N_{i=2}p_\theta(x^i|x^{<i}, c),
\end{equation}
where $c$ is the conditioning text-prompt. Note that each subsequent frame $\hat{x}^{>1}$ is drawn from the distribution of potentially drifted states. By reversing the sampled video $(\hat{x}^{N:2}, x^1)$, we arrive at an example of a corrective video trajectory that provides a ground truth next frame given a set of potential drifted frames as shown in Fig.~\ref{fig:corrective_traj}. 

A key consideration is that these collected trajectories $(\hat{x}^{N:2}, x^1)$ are time-reversed. However, these corrective states are still valid for two primary reasons. First, we rely on the assumption that the manifold of text-conditioned video data is closed under time reversal. For instance, a video of a person walking backwards is just as valid as a video of someone walking forwards. Consider $\mathcal{M}_{\text{data}}$ as the manifold of video sequences corresponding to high-density regions of the distribution $p_{\text{data}}$. We make the assumption that, if $x^{1:N} \in \mathcal{M}_{\text{data}}$, then $x^{N:1} \in \mathcal{M}_{\text{data}}$. Second, to make $(\hat{x}^{N:2}, x^1)$ a valid training sample, we simply need to modify the original text prompt $c$ into $c'$, where the modified text prompt reflects the time-reversed nature of the sample. For example, the text prompt ``a person walking'' would instead be replaced by ``\textit{a reversed video of} a person walking''. 

\paragraph{BAgger Training Loop.} Leveraging these corrective trajectories, we implement the BAgger training loop as an iterative training procedure~\cite{Ross2010ARO}. We begin with an autoregressive video diffusion model $p_{\theta_0}$, pre-trained using a standard Diffusion Forcing~\cite{Chen2024DiffusionFN} objective on a dataset of ground truth video clips $\mathcal{D}{\text{seed}}$. Then, for each round $k$ of BAgger training we: (1) sample a set of drifted video rollouts under the current AR diffusion model, (2) reverse these sampled rollouts and combine them into an aggregated dataset $\mathcal{D}{\text{agg}}$, and (3) train a new model on the aggregated dataset following the original diffusion forcing objective. We formalize this iterative training procedure in Algorithm~\ref{alg:dagger_loop}.

\begin{figure*}[t]
    \centering
    \includegraphics[width=\linewidth]{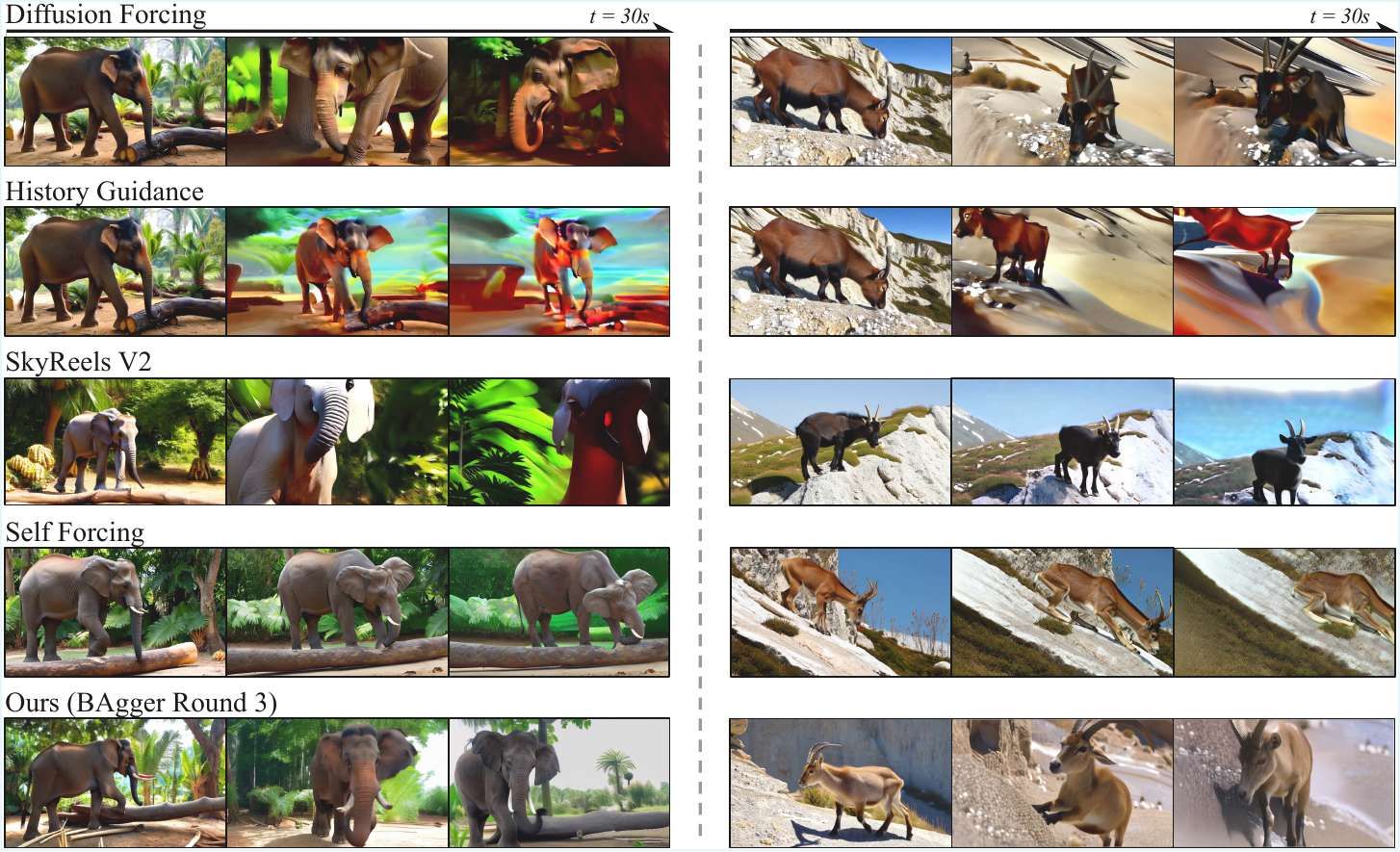}
    \vspace{-2em}
    \caption{\textbf{Qualitative Comparisons.} Side-by-side long-horizon text-to-video results. We compare Diffusion Forcing, History Guidance, SkyReels-V2, Self Forcing, and our BAgger (Round 3). Baseline AR methods exhibit exposure bias artifacts such as over-smoothing, over-saturation, and reduced motion diversity, whereas BAgger better preserves subject identity, scene layout, and local contrast over time. All models share the same architecture with 1.3B parameters. Results show generated frames at 0/15/30s time steps.}
    \label{fig:qualitative_results}
    \vspace{-1.5em}
\end{figure*}

\begin{algorithm}[t!]
\caption{BAgger Algorithm}
\label{alg:dagger_loop}
\begin{algorithmic}[1]
\Require Pre-trained bidirectional diffusion model $p_{\theta}$
\Require Pre-trained AR diffusion model $p_{\theta_0}$
\Require Aggregated dataset $\mathcal{D}_{\text{agg}} \leftarrow \mathcal{D}_{\text{seed}}$
\Require DAgger rounds $K$, Videos per round $M$
\For{$k = 0$ to $K-1$}
    \State $\mathcal{D}_k \leftarrow \emptyset$ \Comment{Initialize dataset for round $k$}
    \For{$m = 1$ to $M$} \Comment{Sample $M$ drifted videos}
        \State Sample $x^1 \sim p_{\text{data}}(x)$, $c$
        \State $(x^1, \hat{x}^{2:N}) \sim \prod^N_{i=2}p_{\theta_k}(x^i|x^{<i}, c)$
        \State $\text{video}_{\text{corr}} \leftarrow (\hat{x}^{N:2}, x^1)$
        \State $c' \leftarrow \text{ModifyPrompt}(c, \text{``in reverse''})$
        \State $\mathcal{D}_k \leftarrow \mathcal{D}_k \cup \{(\text{video}_{\text{corr}}, c')\}$
    \EndFor
    \State $\mathcal{D}_{\text{agg}} \leftarrow \mathcal{D}_{\text{agg}} \cup \mathcal{D}_k$ \Comment{Aggregate corrective traj.}
    \State $p_{\theta_{k+1}} \leftarrow \text{DF}(p_{\theta}, \mathcal{D}_{\text{agg}})$ \Comment{Train on aggredage data}
    \EndFor
\Ensure Optimized model $p_{\theta_K}$
\end{algorithmic}
\end{algorithm}


\paragraph{Architecture.} 
We focus on a diffusion transformer architecture~\cite{Peebles2022ScalableDM} and train our models using the Diffusion Forcing objective (Eq.~\ref{eq:diffusion_forcing}) with a block causal mask~\cite{ai2025MAGI1AV,Yin2024FromSB,Po2025LongContextSV,Huang2025SelfFB}. Data drawn from $\mathcal{D}_{\text{seed}}$ is trained with the DF objective on every frame without modification. Corrective data drawn from $\mathcal{D}_k$ is trained with a modified objective, where a set number of frames are chosen as a prefix. These prefix frames are kept clean when fed as input into the diffusion model, and the resulting losses are not calculated on these frames. We treat these prefix frames as the ``drifted'' states, therefore the model should not learn their distribution, and only learn the distribution conditioned on these states.

\section{Experiments}
\label{sec:experiments}

\paragraph{Implementation.}
We implement BAgger on a pre-trained bi-directional diffusion transformer, namely Wan2.1 1.3B~\cite{Wang2025WanOA}. We modify the model to have block-causal attention~\cite{ai2025MAGI1AV,Yin2024FromSB,Po2025LongContextSV,Huang2025SelfFB} and train using the Diffusion Forcing objective. The model is trained on 5s long videos at 16FPS and $832\times480$ resolution~\cite{Wang2025WanOA}. The model is trained on the latent space produced by a 3D VAE~\cite{Lopez2020DecisionMakingWA, Oord2017NeuralDR}, mapping 81 frames per video to 21 latent frames. We opt for a block-causal mask of 3 latent frames per chunk (total of 7 chunks) for our causal diffusion forcing model. Efficient training is also enabled through the use of FlexAttention~\cite{Dong2024FlexAA}.

\paragraph{Data and Training.}
We build $\mathcal{D}_{\text{seed}}$ from 55K high-quality clips from the Pexels~\cite{pexels} video dataset, with extended captions sourced from MiraData~\cite{Ju2024MiraDataAL}. The initial diffusion forcing model is trained on this dataset for 16K iterations and a batch size of 96. Please refer to the supplementary materials for a detailed discussion of training configurations and hyperparameters.

For each round of BAgger, we generate 27K corrective trajectories (50\% of seed dataset), and directly aggregate all new videos into the next round of training. For each corrective trajectory, we populate the first frame chunk in the model with ground truth frames and perform video extension using the model trained in the last round to generate the 6 leftover chunks. Samples are decoded into pixel space, reversed, then encoded back into the latent space to form corrective trajectories. During training, clean ground-truth clips are trained with standard diffusion objective, where each frame chunk is given independent noise levels. For corrective clips, the first 4 chunks are kept clean, and the DF objective is only calculated against the remaining 3 chunks. For each round of BAgger, the resulting model $p_{\theta_k}$ is fine-tuned from the base Wan2.1 1.3B model for 16K steps.

\paragraph{Long Video Inference.} While the base model only supports up to 5 seconds of video generation, long video generation can be enabled through the use of a sliding window. Prior works have explored the use of a rolling KV-cache~\cite{Huang2025SelfFB, Cui2025SelfForcingTM}. However, without explicitly modifying training to accommodate this inference scheme, it may introduce artifacts due to out-of-distribution KV-cache values~\cite{Huang2025SelfFB}. To ensure this is not a confounding factor during evaluations, we opt to recompute the KV-cache for every sliding window~\cite{ai2025MAGI1AV, Yin2024FromSB}.

\begin{table*}[t]
    \fontsize{9.75pt}{10pt}\selectfont
    \setlength{\tabcolsep}{5pt}
    \renewcommand{\arraystretch}{1.15}
    \begin{tabular}{lcccccccc}
        \toprule
        \multirowcell{2}[-1ex][l]{Method} &
        \multicolumn{6}{c}{\textbf{Global Metrics} $\uparrow$} &
        \multicolumn{2}{c}{\textbf{Drifting Metrics} $\downarrow$} \\
        \cmidrule(lr){2-7} \cmidrule(lr){8-9}
        & Subject & Bg.~Cons. & Smooth. & Dynamic & Aesthetic & Imaging &
          $\Delta^\text{Aesthetic}$ & $\Delta^\text{Imaging}$ \\
        \midrule
        Diffusion Forcing~\cite{Chen2024DiffusionFN} ($\sigma_{\text{test}} = 0$)   & 80.70 & 87.74 & 98.11 & \best{81.25} & 53.05 & \worst{59.97} & 5.12 & \worst{7.34} \\
        Diffusion Forcing~\cite{Chen2024DiffusionFN} ($\sigma_{\text{test}} = 0.2$) & 82.13 & \worst{88.76} & 98.37 & 75.78 & \worst{53.98} & 57.49 & 7.83   & 12.27   \\
        History Guidance~\cite{Song2025HistoryGuidedVD}                                 & 79.15 & 86.20 & \midc{98.65} & 59.38 & 49.24 & 54.33 & 8.83 & 14.56 \\
        \midrule
        \rowcolor[gray]{0.9}
        \multicolumn{9}{l}{\textit{Our Method (Multiple Rounds)}} \\
        BAgger Round 1                            & \midc{82.69} & \midc{89.02} & \best{99.03} & 61.72 & 53.84 & 53.67 & \worst{4.84} & 11.83 \\
        BAgger Round 2                            & \worst{82.29} & 88.70 & 98.49 & \midc{80.47} & \midc{54.68} & \midc{59.98} & \midc{3.79} & \midc{5.92}  \\
        BAgger Round 3                            & \best{84.05} & \best{89.58} & \worst{98.61} & \worst{76.56} & \best{55.35} & \best{63.41} & \best{3.29} & \best{3.57} \\
        \bottomrule
    \end{tabular}
    \vspace{-0.8em}
    \caption{\textbf{Quantitative Comparisons.} VBench evaluation on 50s text-to-video generations averaged over a diverse prompt set. ``Global Metrics'' ($\uparrow$) report subject/background consistency, motion smoothness/dynamic degree, and per-frame aesthetic/imaging quality. Because per-frame means can mask drift, we additionally report ``Drifting Metrics'' ($\downarrow$): the change (first 20\% - average) for aesthetic and imaging scores, lower indicates less degradation. All baselines (Diffusion Forcing with $\sigma_{\text{test}} \in{0,0.2}$, History Guidance) are trained on the same seed data and compute as ours. BAgger improves steadily across rounds, with Round 3 achieving the strongest frame-wise quality and consistency while maintaining comparable motion metrics.}
    \label{tab:finetune_comparisons}

    \vspace{-1em}
\end{table*}

\begin{table}[t]

    \fontsize{7.7pt}{8pt}\selectfont
    \setlength{\tabcolsep}{4pt}
    \renewcommand{\arraystretch}{1.15}

    \begin{tabular}{lccccc}
        \toprule
        \textbf{Model} &
        \shortstack{\textbf{Model}\\\textbf{Params}} &
        \shortstack{\textbf{Teacher}\\\textbf{Model?}} &
        \textbf{NFE} &
        \shortstack{\textbf{Motion}\\\textbf{Quality}} $\uparrow$ &
        \shortstack{\textbf{Frame}\\\textbf{Quality}} $\uparrow$ \\
        \midrule

        MAGI-1~\cite{ai2025MAGI1AV} &
        4.5B &
        No &
        20 &
        \worst{73.11} &
        \worst{57.06} \\

        SkyReels-V2~\cite{Chen2025SkyReelsV2IF} &
        1.3B &
        No &
        20 &
        71.87 &
        56.72 \\

        Self Forcing~\cite{Huang2025SelfFB} &
        1.3B &
        \textbf{\textcolor{red}{Yes (14B)}} &
        4 &
        \midc{73.93} &
        \best{64.81} \\
        \midrule

        Ours &
        1.3B &
        No &
        20 &
        \best{87.59} &
        \midc{59.38} \\
        \bottomrule
    \end{tabular}
    \vspace{-1em}
    \caption{\textbf{Comparison to open-source AR models.} We report parameters, presence of a teacher, denoising steps (NFE), and motion/frame-quality scores. Despite using no teacher, BAgger attains substantially higher motion quality and competitive frame quality (second only to Self Forcing distilled from a 14B teacher).}
    \label{tab:open_comparisons}
    \vspace{-2em}
\end{table}

\subsection{Text-to-Video Generation}

\paragraph{Baselines.} We compare against prior methods for training autoregressive diffusion video models. Specifically, we make comparisons to Diffusion Forcing~\cite{Chen2024DiffusionFN} and History Guidance~\cite{Song2025HistoryGuidedVD}. We compare out method against these baselines trained using the seed dataset and equal number of training iterations as our method. For diffusion forcing, we evaluate on two settings, one where all context frames are kept clean, and the other where a small amount of noise is added to the context frames during generation. For history guidance, we adopt ``fractional history guidance'' with a guidance scale of $w\! =\! 1.2$. We additionally compare our method against open-sourced AR video models. This includes two autoregressive models MAGI-1!\cite{ai2025MAGI1AV} and SkyReels-V2~\cite{Chen2025SkyReelsV2IF}, along with Self Forcing~\cite{Huang2025SelfFB}, a few-step model distilled from Wan2.1 14B. For all multi-step models, we perform inference with 20 denoising steps.

\paragraph{Evaluation Metrics.} We compare our method against relevant baselines on the task of long text-to-video generation. For each method, we sample 50s long videos across a diverse set of text prompts covering human/animal motion, natural landscapes, and more~\cite{Polyak2024MovieGA}. Please refer to the supplementary materials for a detailed discussion on sampling configurations for each baseline.
We adopt VBench~\cite{Huang2023VBenchCB, Huang2024VBenchCA, Zheng2025VBench20AV} as our method of evaluation. Specifically, we evaluate each method along 6 distinct dimensions, which includes subject and background consistency, the smoothness and dynamic degree of motion, and the per-frame aesthetic and imaging quality. The per-frame quality metrics for each video are calculated as an average across all frames, therefore it struggles to capture the decay in quality caused by exposure bias. With this in mind, following prior work~\cite{Zhang2025FrameCP}, we additionally report the change in imaging and aesthetic quality over the course of each long video by taking the difference in scores between the first 20\% of frames and the average across all frames in each generated video.

\paragraph{Qualitative results.} Figure~\ref{fig:qualitative_results} presents side-by-side text-to-video comparisons. BAgger maintains subject identity, scene layout, and local contrast over substantially longer horizons than baseline AR methods. While exposure bias manifests differently across approaches, common failure modes include over-smoothing of foreground and background details, over-saturation, and reduced motion diversity. Diffusion Forcing exhibits pronounced over-smoothing and over-saturation, with SkyReels-V2 showing the same tendencies. History Guidance, by tightly anchoring to context frames, over-attends to already drifted states, further intensifying saturation and suppressing motion diversity in later segments. Self Forcing delivers strong initial frame quality (benefiting from a 14B teacher) but progressively loses dynamics, with error accumulation leading to near-static later frames alongside with saturation artifacts also settling in. In contrast, BAgger’s trajectories remain visually stable and motion-rich over time.

\paragraph{Quantitative comparisons.} We report quantitative comparisons with Diffusion Forcing and History Guidance, trained on the same dataset under a matched compute budget, in Tab.~\ref{tab:finetune_comparisons}. After three BAgger rounds, our model attains the best frame-wise quality and subject/background consistency, while maintaining comparable performance on motion-related metrics. We further benchmark against open-source baselines in Tab.~\ref{tab:open_comparisons}. For readability, we group metrics into \emph{motion-related} and \emph{frame-quality} categories. Relative to open-source models, our method yields higher motion quality and stronger per-frame scores than other non-distilled methods, trailing only Self Forcing, which leverages distillation from a 14B teacher to achieve higher frame-wise quality. While Self Forcing produces videos with great frame-wise quality, it achieves this by generating relatively static video sequences. This lack in motion diversity is reflected in its relatively poor motion quality scores.
\subsection{Applications.}
\paragraph{Multi-Prompt Generation.} 
A natural application of our method lies in multi-prompt video generation~\cite{Yin2024FromSB}, where the textual conditioning can evolve dynamically throughout the generation process. Unlike conventional text-to-video models that rely on a single static prompt~\cite{Wang2025WanOA, videoworldsimulators2024, Yang2024CogVideoXTD}, our approach allows prompt changes on the fly during autoregressive inference. This enables the model to transition smoothly between distinct scenes or actions while maintaining temporal consistency. As illustrated in Fig.~\ref{fig:main_results}, this capability allows the creation of long, coherent videos with fine-grained and diverse control over content and motion through dynamic prompting. Such flexibility has broad implications for world modeling~\cite{oasis2024,genie3,parkerholder2024genie2,Bruce2024GenieGI,Valevski2024DiffusionMA} and interactive video generation~\cite{Shin2025MotionStreamRV, Huang2025SelfFB}, where user inputs or simulated environmental events can continuously steer the evolving visual narrative in real time.

\paragraph{Video Extension.}
As a result of diffusion forcing training, our model naturally supports video extension by initializing the key–value cache of the first chunk with embeddings from an input clip, enabling seamless continuation. While our block-causal attention pattern is designed for multi-frame chunks and does not directly support image-to-video, this can be achieved by modifying the causal pattern to include a single latent frame in the first chunk, as done in prior work~\cite{Huang2025SelfFB,Shin2025MotionStreamRV}. As shown in Fig.~\ref{fig:main_results}, our model produces smooth and consistent extensions, demonstrating strong generalization to continuation tasks.

\subsection{Ablation}
\begin{figure}[t]
    \centering
    \includegraphics[width=\linewidth]{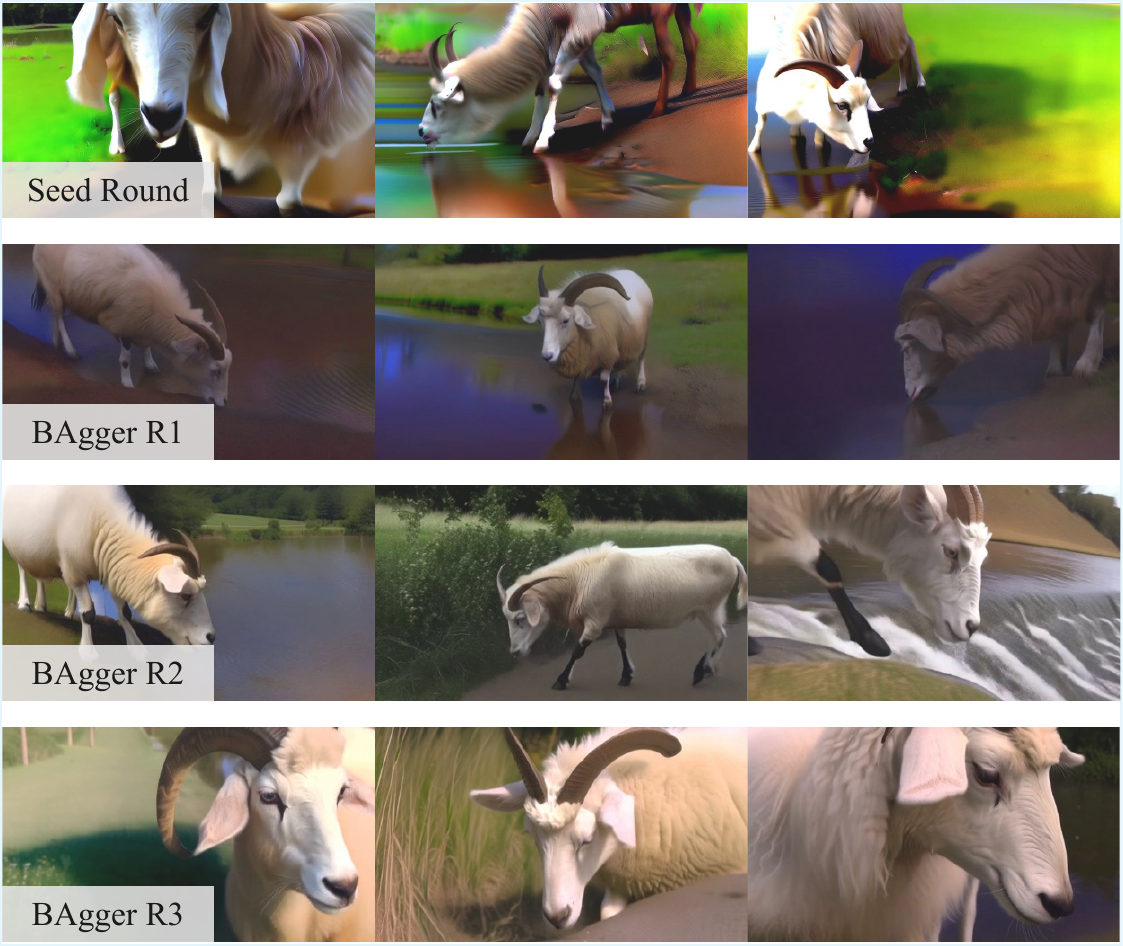}
    \vspace{-1.8em}
    \caption{\textbf{Ablation Over BAgger Rounds.} Final frames from three 30-second generations per model. Training on seed data alone leads to severe over-saturation, while one round of BAgger over-corrects toward under-saturated outputs. By the second round, color balance and contrast stabilize, and the third round further improves visual fidelity and temporal consistency.}
    \label{fig:ablation}
    \vspace{-1.5em}
\end{figure}

\paragraph{Effect over Multiple BAgger Rounds.}
We ablate across multiple rounds of the BAgger algorithm to study its effect on long-horizon generation. A single round is insufficient to capture the full distribution of drifted states and can even degrade performance relative to training on seed data alone. As shown in Tab.~\ref{tab:finetune_comparisons}, performance improves steadily over multiple rounds. Fig.~\ref{fig:ablation} visualizes examples from each round, sampling three 30-second generations per model and showing the final frame for each sample. The seed-only model exhibits severe over-saturation, while one round of BAgger over-corrects toward under-saturated outputs. By the second round, saturation and contrast stabilize, producing more natural color balance, and the third round yields further improvements in detail and temporal consistency.

\section{Discussion}
\label{sec:discussion}
\paragraph{Limitations and Future Work.} 
While our method effectively mitigates drift, drifting artifacts can still appear over very long horizons, as a few rounds of BAgger may not fully cover the space of drifted states. For fair comparison, we retrain each round from the bidirectional teacher to isolate the effect of aggregation rather than accumulated compute. However, as shown in prior work~\cite{Xu2025CompliantRD}, continuing training from the previous round could further improve performance and efficiency. The ratio between seed and corrective data also remains an open design choice that may influence stability and generalization. Finally, although inference at 20 steps makes KV-cache re-initialization overhead negligible, future work could explore training with rolling KV-cache updates~\cite{Cui2025SelfForcingTM, Shin2025MotionStreamRV, Cui2025SelfForcingTM} to improve runtime efficiency. While out model is not real-time, future work could explore layering few-step distillation on top of BAgger. Crucially, since BAgger does not rely on distribution-matching losses, distillation would not be constrained to mode-seeking objectives used in prior works~\cite{Huang2025SelfFB, Yin2024FromSB}.

\paragraph{Conclusion.} We present Backwards Aggregation (BAgger), a self-supervised approach to mitigating exposure bias in autoregressive video diffusion. By reversing a model's own drifting rollouts to form corrective trajectories, BAgger gradually trains on the model's own induced distribution, eliminating the train-test mismatch at the heart of autoregressive deterioration. Implemented with a causal diffusion transformer, it improves long-horizon generation without modifications to the original flow or score-matching objective. Crucially, BAgger avoids distribution-matching losses used in prior work~\cite{Yin2024FromSB,Huang2025SelfFB}, helping preserve motion diversity without sacrificing visual fidelity, as supported by our qualitative and quantitative evaluations.

{
    \small
    \bibliographystyle{ieeenat_fullname}
    \bibliography{main}
}

\clearpage
\setcounter{page}{1}
\maketitlesupplementary
\setcounter{section}{0}
\renewcommand{\thesection}{\Alph{section}}
\renewcommand{\thesubsection}{\thesection.\arabic{subsection}}

\section{Video Results}
\label{sec:details}
\paragraph{Supplementary Website.} Because static figures in the main manuscript cannot fully convey our video results, we provide corresponding videos for each main result on the supplementary website. These can be accessed via the attached HTML file. Due to the 200MB submission size limit and the length of our videos, many clips have been compressed and may appear lower quality than the original generations.

\section{Implementation Details}
\label{sec:details}

\subsection{Training Configurations}
\paragraph{Model.} We primarily build our method upon the Wan2.1~\cite{Wang2025WanOA} and Self Forcing~\cite{Huang2025SelfFB} codebases. We primarily work with the Wan2.1 1.3B Text-to-Video model as our base model, fine-tuning it into a block-causal model using FlexAttention~\cite{Dong2024FlexAA}. As Wan2.1 1.3B supports bidirectional generation of 21 latent frames, we follow prior works~\cite{Yin2024FromSB, Huang2025SelfFB} and use a chunk size of 3 frames per chunk, resulting in 7 total chunks. This configuration provides a reasonable balance between parallelism and causal granularity. We apply the same chunking and causal masking scheme to all autoregressive baselines to ensure a fair architectural comparison.

\paragraph{Data.}We use a subset of the Pexels video dataset~\cite{pexels} for causal fine-tuning and generation of corrective trajectories. Specifically, we use the Pexels subset from the MiraData~\cite{Ju2024MiraDataAL} dataset, and use the ``dense prompt'' field as the corresponding text for each corresponding video clip. Due to varying FPS and clip lengths in the dataset, we subsample videos that have higher than 16FPS to match the frame rate of our video model (16FPS) and randomly sample a 5 seconds clip from each data sample. For generation of corrective trajectories, we take the initial 9 frames = 3 latent frames = 1 chunk, from a chosen video clip and perform rollouts conditioned on an initial ground truth chunk and the corresponding text prompt.

\paragraph{Compute and Hyperparameters.} We perform all of our experiments on 16 NVIDIA H100 GPUs (80GB VRAM per GPU). For each round of training, we perform 16,000 training iterations at a batch size of 96, using a constant learning rate of $4e^{-6}$. We follow the original Wan2.1 flow matching~\cite{Lipman2022FlowMF} objective, and adopt a time step shift in the form, $t'(k, t) = (kt/1000)/(1 + (k -1)(t/1000))\cdot1000$ and a shift factor $k = 8$.

\subsection{Inference Configurations}
We perform inference for all multi-step AR diffusion models (ours and baselines) at 20 inference steps per chunk. For our model, we perform inference with a classifier-free guidance scale of $w = 6$, and a time step shift coefficient of $k=8$. For corrective trajectory generations, the same inference configurations are used. For long video generations with sliding windows, we perform KV-cache re-initialization at every sliding window step to eliminate any potential issues caused by OOD KV-cache values as mentioned in Xun et al.~\cite{Huang2025SelfFB}. Each sliding window instance, we keep 12 latent frames from the tail end of the previous generation as the initial context of the next set of generations. For multi-prompt generations, we simply change the text-prompt conditioning during the sliding window change.
\begin{table}[t]
    \fontsize{9pt}{10pt}\selectfont
    \setlength{\tabcolsep}{3.6pt}
    \renewcommand{\arraystretch}{1.15}

    \begin{tabular}{lcccc}
        \toprule
        \textbf{Model} &
        \textbf{Smooth.} &
        \textbf{Dynamic} &
        \textbf{Aesthetic} &
        \textbf{Imaging} \\
        \midrule

        MAGI-1~\cite{ai2025MAGI1AV} &
        99.35 & 46.88 &	55.69 &	58.43 \\
        SkyReels-V2~\cite{Chen2025SkyReelsV2IF} &
        99.20 & 44.53 & 55.60 & 57.84\\

        Self Forcing~\cite{Huang2025SelfFB} &
        98.64 & 49.22 & 58.33 & 71.29 \\
        \midrule

        Ours &
        98.61 & 76.56 & 55.35 & 63.41 \\
        \bottomrule
    \end{tabular}
    \vspace{-0.8em}
    \caption{Extended results comparing our method against relevant open baseline models. Our model shows much better motion dynamics while maintaining simialr levels of motion smoothness. For per-frame quality metrics, on average, our method is only second to Self Forcing which is distilled from a 14B base model. Our method outperforms other AR diffusion models (MAGI-1 and SkyReels) in imaging quality, and has similar aesthetic performance, despite being fine-tuned on a much smaller dataset.}
    \label{tab:full_open}
    \vspace{-1em}
\end{table}
\begin{figure*}[t]
    \centering
    \includegraphics[width=\linewidth]{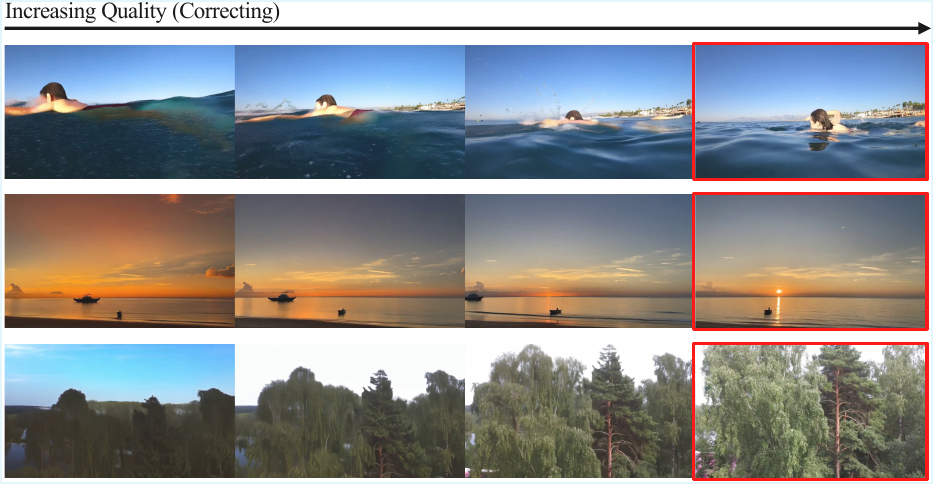}
    \vspace{-1.8em}
    \caption{More examples of corrective trajectories gathered from first round of BAgger. Ground truth frames are highlighted in red. Note that the above sequences show generated frames in reverse order, therefore the order above is the order in which the model is trained on.}
    \vspace{-1em}
    \label{fig:corrective_traj}
\end{figure*}
\section{Evaluations}

\subsection{Baseline Details}
\paragraph{History Guidance.} We implement a version of History Guidance resembling Song et al.~\cite{Song2025HistoryGuidedVD}, performing joint guidance between the text-conditioned prediction and fractional history guidance as defined in ~\cite{Song2025HistoryGuidedVD}. The resulting velocity prediction used takes the form,
\begin{equation}
\begin{aligned}
    v_{\text{hg}}(x^{i}_t, t, c) 
    &= v(x^{<i}, t, \emptyset) \\
    &\quad +\, w_{\text{text}} \big[\, v(x^{<i}, t, c) - v(x^{<i}, t, \emptyset) \,\big] \\
    &\quad +\, w_{\text{hg}} \big[ v(x^{<i}_p, t, \emptyset) - v(\emptyset^{<i}, t, \emptyset)  \big].
\end{aligned}
\end{equation}
Where $w_{\text{text}}$, and $w_{\text{hg}}$ are the guidance scale for the text conditioned and history conditioned terms, $p = 800$ is a partial noise level chosen for fractional history guidance, $c$ is the text prompt, and $\emptyset$ represents the empty string, and $\emptyset^{<i}$ randomly sampled noise in the same shape as the context. In our evaluations we use $w_{\text{text}} = 6$ and $w_{\text{hg}} = 1.2$.

\subsection{Evaluations}

\paragraph{Metric Details.} We base our evaluations on the custom evaluation suite of VBench~\cite{Huang2023VBenchCB,Huang2024VBenchCA,Zheng2025VBench20AV}, which spans the 6 dimensions. As mentioned for Tab.~\ref{tab:finetune_comparisons}, we also report drifting metrics which measures the drop in per-frame quality metrics between the first 20\% of all frames and the average across all frames. We also report the full metrics for Tab.~\ref{tab:open_comparisons} in Tab.~\ref{tab:full_open}. Note that the results in Tab.~\ref{tab:open_comparisons} were calculated by taking the averages between the two motion metrics, and two frame-wise quality metrics respectively.

\section{Additional Results}

\paragraph{Rollout Examples.} We provide additional examples of corrective/drifting trajectories generated during the BAgger algorithm. Our model takes in a set of ground truth frames and generated a set of (potentially) drifted extensions of these frames by rolling out the AR model trained from the previous round.

\section{Ethical Considerations.}
Video generative models inherently carry substantial potential for misuse. High-fidelity outputs can enable convincing deepfakes that are increasingly difficult to distinguish from real footage, posing risks for misinformation, manipulation, and privacy. Without careful governance and responsible deployment, these models may also amplify harmful societal biases present in their training data. At the same time, they offer meaningful benefits, such as powering world models for robotics, enhancing accessibility, and supporting safer real-world decision-making. We acknowledge this duality and emphasize the importance of continued research, policy development, and safeguards that mitigate negative impacts while enabling socially beneficial uses.

\end{document}